\pdfoutput=1

\documentclass[11pt]{article}

\usepackage[]{EMNLP2022}

\usepackage{times}
\usepackage{latexsym}

\usepackage[T1]{fontenc}

\usepackage[utf8]{inputenc}

\usepackage{microtype}
\usepackage{inconsolata}
\usepackage{balance}
\usepackage[flushleft]{threeparttable}
\usepackage{enumitem}
\usepackage[linesnumbered,ruled,vlined]{algorithm2e}
\usepackage{algpseudocode}
\usepackage{booktabs}
\usepackage{subcaption}
\usepackage[nointegrals]{wasysym}
\usepackage{tikz}
\usepackage{tabu}
\usepackage{tabularx}
\usepackage{float}
\usepackage{listings}
\usepackage{color}
\usepackage{multirow}
\usepackage{framed}
\usepackage{pifont}
\usepackage{dsfont}
\usepackage{xcolor}
\usepackage{amsmath,amsthm}
\usepackage{amssymb}
\usepackage{graphicx}
\usepackage{textcomp}
\usepackage{tcolorbox}
\usepackage{soul}
\usepackage{makecell}
\usepackage[normalem]{ulem}
\usepackage{xspace}
\usepackage{arydshln}

\newcommand{\ruleie}{\textsc{RuleSI}}
\newcommand{\ours}{\textsc{Dissi}}
\newcommand{\base}{\textsc{SI}}
\newcommand{\keywordCode}[1]{{\small \texttt{#1}}}

\title{Cross-Lingual Speaker Identification Using Distant Supervision}

\author{
  Ben Zhou\textsuperscript{1\thanks{~~Most work done while interning at Tencent AI Lab.}} ~~ Dian Yu\textsuperscript{2} ~~ Dong Yu\textsuperscript{2} ~~ Dan Roth\textsuperscript{1} \\
   \textsuperscript{1} University of Pennsylvania, Philadelphia, PA \\
   \textsuperscript{2}Tencent AI Lab, Bellevue, WA \\
\{xyzhou, danroth\}@seas.upenn.edu, \{yudian, dyu\}@tencent.com
}

\begin{document}
\maketitle
\begin{abstract}
Speaker identification, determining which character said each utterance in literary text, benefits many downstream tasks. Most existing approaches use expert-defined rules or rule-based features to directly approach this task, but these approaches come 
with significant drawbacks, such as 
lack of contextual reasoning and poor cross-lingual generalization.
In this work, we propose a speaker identification framework that addresses these issues. We first extract large-scale distant supervision signals in English via general-purpose tools and heuristics, and then apply these weakly-labeled instances with a focus on encouraging contextual reasoning to train a cross-lingual language model. We show that the resulting model outperforms previous state-of-the-art methods on two English speaker identification benchmarks by up to $9\%$ in accuracy and $5\%$ with only distant supervision, as well as two Chinese speaker identification datasets by up to $4.7\%$.





\end{abstract}

\section{Introduction}
\label{sec:intro}

Speaker identification (SI) aims to decide who said or implied each quote/utterance in a document~\cite{elson2010automatic}. It is mostly studied in the domain of literature and novels because, unlike news, the speakers in stories are often not explicitly specified by a name. Thus, contextual reasoning is required for most SI tasks~\cite{Jia2020SpeakerIA} (e.g., the implicit speaker of paragraph $4$ (P$4$) is \emph{``Wickham''} in Table~\ref{table:example}). Besides, as SI datasets are usually too small-scale to effectively train large pre-trained language models, most previous studies boost the mono-lingual performance by additionally designing language-specific patterns and heuristics, which require mono-lingual expertise and cannot be easily transferred to other languages.



\begin{figure}[t!]
    \centering
    \includegraphics[width=0.45\textwidth]{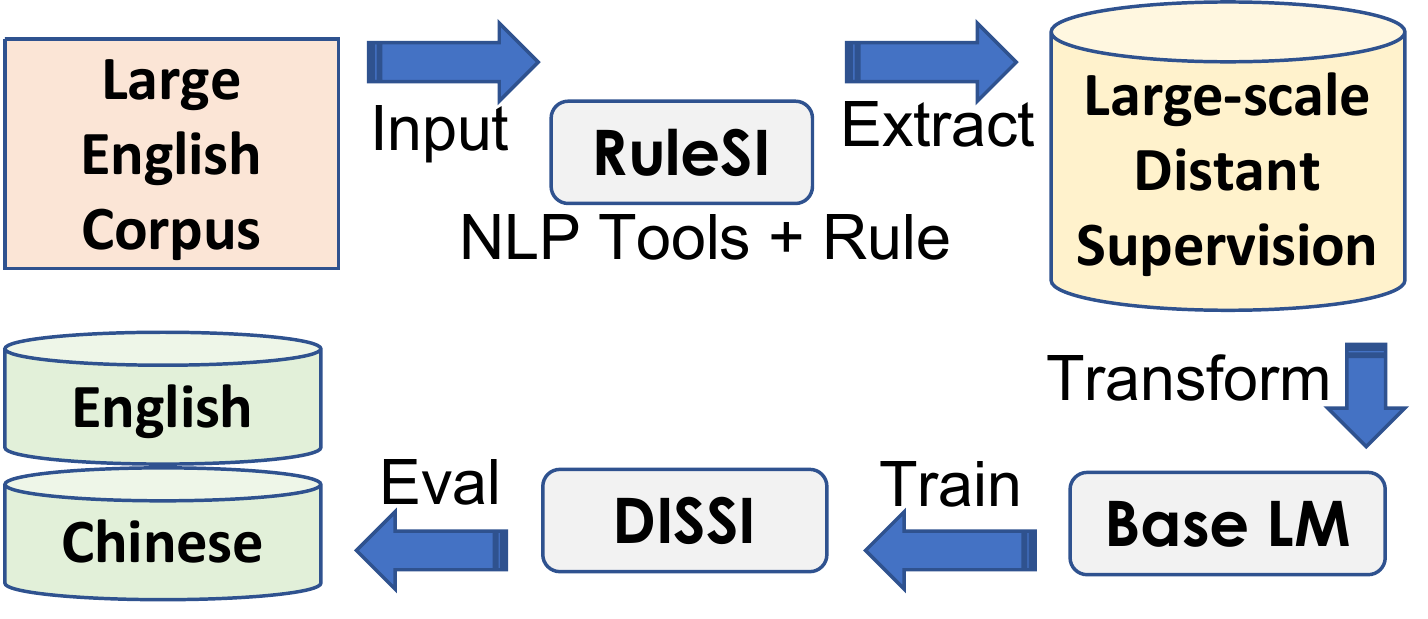}
    \vspace*{-1mm}
    \caption{Overview of our framework. \ruleie{} extracts incidental supervision signals used to train \ours{}.}
    \vspace*{-1mm}
    \label{fig:overview}
\end{figure}

\begin{table}[h!]
\centering
\scriptsize
\begin{tabular}{p{0.1cm}p{6.9cm}}
\midrule
ID & Content \\
\midrule
P1& \emph{The contents of this letter threw \textbf{Elizabeth} into a flutter of spirits ...}\\
P2& \emph{... she was overtaken by \textbf{Wickham}...} \\
P3& \emph{``You certainly do,'' \underline{she} replied with a smile ...}\\
P4& \emph{``I should be sorry indeed, if it were.  We were always good friends; and now we are better.''}\\
P5& \emph{``True.  Are the others coming out?''} \\
\bottomrule
\end{tabular}
 \caption{An example requiring contextual reasoning from the P\&P dataset~\cite{He2013IdentificationOS}.}
 \label{table:example}
\end{table}


In this work, we address these issues with a novel framework for cross-lingual SI \textit{without relying} on any domain, task, or language-specific annotation for a new language. The framework, as overviewed in Fig.~\ref{fig:overview}, starts with extracting large-scale distant and incidental supervision \cite{Roth2017IncidentalSM} from unstructured English corpora. \textbf{We propose a rule-based system} called \ruleie~for extraction (\S\ref{sec:mono-lingual-ie}). \textbf{We collect large-scale (55K) weakly-labeled English instances} with \ruleie{} to enable the training of large, advanced models and transform them to encourage more contextual reasoning (\S\ref{sec:distant-supervision}). \textbf{We train the first cross-lingual SI model} with the constructed data and call the resulting model \textbf{\ours{}} (\textbf{Di}stantly-\textbf{S}upervised \textbf{S}peaker \textbf{I}dentification). We hypothesize that \ours{} can improve cross-lingual performance because the SI task shares many language-invariant features (\S\ref{sec:cross-lingual}). 

Experimental results\footnote{We release code and data at: \url{https://github.com/Slash0BZ/speaker-identification}.} show that \ours{} achieves state-of-the-art English performance on the P\&P dataset~\cite{He2013IdentificationOS}, improving $7.0\%$ in the zero-shot setting and $6.2\%$ with full supervision. With almost no language-specific efforts, our cross-lingual model outperforms state-of-the-art methods on two Chinese datasets WP~\cite{Chen2019ACD,chen2021neural} and JY~\cite{Jia2020SpeakerIA}, by up to $4.7\%$. Comparing to the baseline, our distant supervision brings an improvement of more than $40\%$ in realistic few-shot settings. In particular, \ours{} can be well applied across languages even without any annotation, e.g., achieving 90.6\% zero-shot accuracy on P\&P and 89.5\% on the Chinese JY dataset.





\section{Related Work}
\noindent
\textbf{Speaker Identification.}
Language-specific expert-designed rules, patterns, and features~\cite{elson2010automatic,He2013IdentificationOS,muzny2017two,ek-2018-identifying} are widely used to identify speakers. 
\citet{pavllo2018quootstrap} aim to find and bootstrap over lexical patterns for SI, whereas we focus on using high-precision heuristics to construct distant instances. 
Previous cross-lingual SI studies mainly focus on direct speech identification~\cite{kurfali2020zero,byszuk2020detecting}. To the best of our knowledge, this is the \textbf{first work on cross-lingual SI} without the need for redesigning rules, patterns, and features for a new language.

\noindent
\textbf{Indirect Supervision.} Studies have shown that distant or indirect supervision is effective in bridging the knowledge gaps in pre-trained language models (LMs)~\cite{Zhou2020TemporalCS, Zhou2021TemporalRO, Khashabi2020UnifiedQACF}. ~\newcite{yu-etal-2022-end} improve the SI performance with self-training while a large-scale clean dataset is required for training teacher models.


\section{English Speaker Identification}
\label{sec:mono-lingual-ie}
In this section, we introduce a rule-based SI system named \ruleie{} (\textbf{Rule}-based \textbf{SI}): it receives a long document as input and then outputs (context, utterance, speaker) tuples from the document. \ruleie{} can be directly applied to identify speakers in English texts in a given dataset, but we mainly use it\footnote{This is because \ruleie{} is not guaranteed to produce a predicted speaker for every utterance due to pattern coverage.} to automatically extract incidental signals that approximate the target task from unlabeled corpora, later used as distant supervision to train our cross-lingual SI system \textbf{\ours{}} in \S\ref{sec:cross-lingual}. 


\subsection{Main Heuristics}
\label{sec:main_heuristics}
\ruleie{} extracts quoted utterances from segmented sentences by simply matching quotation marks. For each extracted utterances, we form a context with its previous three and next two sentences, and find all person characters with a named entity recognition (NER) tool in AllenNLP~\cite{Gardner2017AllenNLP}. In the same context, any name that is a sub-string of a longer name will be merged as the same character. We then employ three heuristics to try to identify a speaker among the characters for each utterance. 
The first two are commonly used rules proposed by \citet{He2013IdentificationOS, muzny2017two}, namely \textit{Direct Speaker Identification} and \textit{Conversation Alternation Patterns}. We follow the same implementation as \citet{muzny2017two}, except that we use an SRL model from AllenNLP to replace dependency parsers. We refer readers to this work for details of first two rules due to space limitations. The first heuristic collects a list of speech verbs (e.g., \emph{``say''}) and uses a dependency parser to find if there is a speech verb connecting a noun phrase and a target utterance. If so, we regard the noun phrase as the speaker of the target utterance. The second heuristic assumes that conversations in novels follow simple speaker alternation patterns. For example, in consecutive utterances in Table~\ref{table:example}, once we identify that the speaker of P$3$ is \emph{``Elizabeth''}, we assume that she is very likely to say the utterance of P$5$. Besides these two rules, we introduce a new heuristic based on coreference to address anaphoric speakers such as \emph{``she''} in P$3$.

\noindent
\textbf{Local Coreference Resolution with Pronouns.}
Previous work~\cite{muzny2017two} use coreference resolution (coref) only for explicit speakers.
We extend the application of coref to all pronouns in the \textbf{utterances}, because i) any character mention that corefs with a first-person pronoun (e.g., \textit{``I''} and \textit{``me''}) inside the utterance reveals the speaker and ii) those that coref with second and third-person pronouns (e.g., \textit{``you''} and \textit{``she''}, ) should be excluded from candidate speakers. We run the AllenNLP coref model on every three-sentence window as coref models that though perform reasonably well on short literal texts often mistakenly reduces the number of clusters in a lengthy text. 
\noindent \textbf{Soft Inference.} All the three above-mentioned rule-based heuristics will assign speakers separately, and they can conflict with each other. As there is no hierarchy among the heuristics, we employ soft assignments by letting each rules to ``vote'' or ``vote against'' for a candidate. We assign the speaker with the highest vote count to each utterance.

\section{Distant Supervision Acquisition}
\label{sec:distant-supervision}
We hypothesize that the SI task shares many commonalities across languages (e.g., the patterns people use to describe different types of speakers in texts). The high-quality data in a resource-rich language may help SI in other languages. This section describes how we use \ruleie{} to acquire large-scale English SI instances for distant supervision. 

\subsection{Automatic Extraction}
We use the Project Gutenberg, which contains over 60,000 books, as the source corpus.\footnote{https://www.gutenberg.org/ (books are not protected by copyright laws and distributed for free use).} 
We run \ruleie{} on the raw texts from these books, which automatically identifies quoted utterances, candidate speakers from nearby contexts, and tries to assign a speaker from the candidates to each utterance. Naturally, not all utterances can be parsed with the high-precision heuristics we use in \S\ref{sec:main_heuristics}.
As a result, we extract 55K (context, utterance, speaker) instances. We view these instances as distant supervision as they are automatically constructed (therefore with a certain level of noise) from external resources and do not rely on any task or domain-specific annotation. 


\subsection{Contextual Reasoning with Masking}
\label{sec:masking}
As argued in \S\ref{sec:intro}, we aim to build models that approach SI with contextual reasoning. However, many of automatically extracted instances have explicit speakers ($53\%$ from the analysis in $\S$\ref{sec:quality_of_weak}) and do not contribute much to a stronger reasoning model. As an improvement, we mask explicit speaker mentions with \textit{``someone''} with a probability of $85\%$, so models are forced to use other textual clues to identify the speaker, which often times involve context-level understanding.
Besides, to avoid the model overfitting to speaker names, which are relatively irrelevant in determining who said each utterance, we randomly assign each character a masked name \emph{``Person [X]''} (where [X] is a random alphabetic letter, but we remove letters that are meaningful by itself such as ``I'' from consideration), and we replace corresponding mentions in the input context with the masked name.

\section{Cross-Lingual Model Formulation}
\label{sec:cross-lingual}
Given the large amount of distant supervision in English, we explore the possibility of transferring mono-lingual signals to cross-lingual applications, under the help of pre-trained cross-lingual LMs. In this section, we propose and describe \ours{}.

We follow very recent SI work~\cite{yu-etal-2022-end} to formulate the task as a classical span selection task~\cite{devlin-2019-bert}: given an input sequence containing the list of candidate characters (\keywordCode{People:[C$_1$][C$_2$]...[C$_n$]}), \keywordCode{context} that contains the target utterance \keywordCode{[U]} and its surrounding sentences, and a corresponding question ``\keywordCode{who said [U]?}'', the task is to extract the speaker of \keywordCode{[U]} from the candidate characters (\keywordCode{[C$_1$]...[C$_n$]}). We adopt a training objective to maximize the sum of the log-likelihoods of the independently predicted start and end positions of the correct speaker. Different from~\cite{yu-etal-2022-end}, we follow the traditional SI setting assuming that candidate speakers are given.


\section{Experiments}
\label{sec:experiments}

\subsection{Data and Baselines}
For English, we use Pride \& Prejudice (P\&P)~\cite{He2013IdentificationOS}. We also report results on the Emma dataset \cite{muzny2017two}, but we remove $127$ test instances due to conflicting aliases (annotation errors), hence making the comparison on Emma with previous work indirect. Since Emma does not provide training data, no in-domain numbers are reported. For Chinese, we use two datasets, one based on \textit{Jinyong} novels (JY) and another based on novel \textit{World of Plainness} (WP). We compare the performance of \ours{} with published best results on each dataset and that of the baseline language model in multiple settings.  We use the provided candidate speakers in these datasets as \keywordCode{[C$_1$]...[C$_n$]} in the input. Note that we replace character mentions with masked names following \S\ref{sec:masking}. More details about data and format are in Appendix~\ref{sec:appendix:data}.

\subsection{Implementation Details}
We use RoBERTa-large \cite{Liu2019RoBERTaAR} for English and XLMR-large \cite{Conneau2020UnsupervisedCR} for other languages such as Chinese as the backbone LMs of baselines, which we denote as \base{}$_{\text{en}}$ and \base{}$_{\text{x}}$, respectively.
Both LMs are trained on our English distant supervision data (Section~\ref{sec:distant-supervision}) for one epoch, and we denote these distant-supervised models as \ours{}$_{\text{en}}$ and \ours{}$_{\text{x}}$, respectively. We use Transformers~\cite{Wolf2020TransformersSN} for the span selection implementation with a learning rate of 3e-5. Both runs finish in an hour with a single RTX A6000. For all experiments with additional in-domain supervision, we run with three random seeds, each time for three epochs with the same learning rate.%

\noindent
\textbf{Inference.}
For English evaluation, we apply an additional inference process to both the baseline LM and our models with \ruleie{}. We treat any \textit{named} mentions identified as direct speakers by \ruleie{} as final predictions. If the direct speaker is a pronoun that indicates genders (e.g., \textit{``he''} and \textit{``she''}), we remove all candidates of other genders from the candidate speakers. 



\subsection{Main Results}


\begin{table}[t]
\centering
\footnotesize
\begin{tabular}{lp{2cm}cc}
\toprule
System & Supervision & \textit{P\&P} & \textit{Emma} \\
\cmidrule(lr){1-1}\cmidrule(lr){2-2}\cmidrule(lr){3-3} \cmidrule(lr){4-4}
\citet{muzny2017two}& no & 83.6 & (75.3) \\
\citet{muzny2017two} & in-domain & 85.2 & -- \\
\midrule
\base$_{\text{en}}$ & in-domain & 71.1 & -- \\ 
\ours{}$_{\text{en}}$ & distant & 90.6 & \textbf{84.7} \\
\ours{}$_{\text{en}}$ & distant & \textbf{91.4} & -- \\
          & +in-domain &     &  \\
\midrule
\ours{}$_{\text{en}}$ - masking & distant & 85.2 & 81.1 \\
\ours{}$_{\text{en}}$ - coref & distant & 85.2 & 82.2 \\
\bottomrule
\end{tabular}
\caption{Accuracy ($\%$) on English SI datasets, with ablation results. Numbers in parentheses are for reference only. \ours{}-* are from this work.}
\label{tab:english}
\end{table}

\begin{table}[t]
\centering
\footnotesize
\begin{tabular}{llcc}
\toprule
System & Supervision & \textit{JY} & \textit{WP} \\
\cmidrule(lr){1-1}\cmidrule(lr){2-2}\cmidrule(lr){3-3}\cmidrule(lr){4-4}
Random$^\dag$ & no  & 33.7 & 37.6 \\
MLP$^\dag$  & in-domain     & 95.6   & 70.5      \\
CSN$^\dag$  & in-domain            & --  & 82.5      \\
E2E\_SI$^\dag$ & in-domain & 98.3 & 80.9 \\
\midrule
\ours{}$_{\text{x}}$ & distant & \textbf{89.5} & \textbf{63.8} \\
\midrule
\base{}$_{\text{x}}$ & mini & 51.7 & 40.9 \\
\ours{}$_{\text{x}}$ & mini+distant & \textbf{95.1} & \textbf{77.5} \\
\midrule
\base{}$_{\text{x}}$ & in-domain & 98.3 & 53.4 \\
\ours{}$_{\text{x}}$ & in-domain+distant & \textbf{98.5} & \textbf{87.2} \\
\bottomrule
\end{tabular}
\caption{Accuracy ($\%$) on Chinese SI datasets ($^\dag$: from previous work~\cite{Jia2020SpeakerIA,chen2021neural,yu-etal-2022-end}, and \textit{mini} uses 200 in-domain instances). }
\label{tab:chinese}
\end{table}

Table~\ref{tab:english} compares English SI accuracy with state-of-the-art (SOTA) numbers \cite{muzny2017two, Yoder2021FanfictionNLPAT}. \ours{}$_{\text{en}}$ outperforms previous SOTA results by $6.2\%$. Even without direct supervision, our model outperforms previous supervised system by $5.4\%$. Comparing to the baseline \base{}$_{\text{en}}$, the same backbone model trained on distant supervision improves over $20\%$. 

Table~\ref{tab:chinese} shows performance on Chinese benchmarks. With full supervision, our model \ours{}$_{\text{x}}$ improves $2.9\%$ and $4.7\%$ over previous SOTA on JY and WP respectively, and gains $34\%$ over the same baseline without distant supervision on WP. We also achieve comparable performance (+$43\%$) on JY with only $1\%$ of training instances (\textit{mini}). 

As Table~\ref{tab:chinese-category} shows, we find that our method outperforms previous methods on identifying all three types of speakers by a large margin. On the WP dataset that provides ground truth type labels for instances, for the most challenging implicit category, our method obtains a $8\%$ improvement compared with the previous SOTA performance.

\begin{table}[t]
\centering
\footnotesize
\begin{tabular}{lccc}
\toprule
System & Explicit & Anaphoric & Implicit \\
\cmidrule(lr){1-1}\cmidrule(lr){2-2}\cmidrule(lr){3-3}\cmidrule(lr){4-4}
\base{}$_{\text{x}}$        &  52.3     &  54.2 &  48.3  \\
CSN  &  93.2      & 81.3  & 75.9    \\
\ours{}$_{\text{x}}$    & \textbf{100.0}    & \textbf{85.2} & \textbf{83.9} \\
\bottomrule
\end{tabular}
\caption{Accuracy (\%) by type according to the WP dataset. Results are produced with full supervision.}
\label{tab:chinese-category}
\end{table}













\subsection{Ablation Studies and Analysis}
\label{sec:quality_of_weak}
Table~\ref{tab:english} also includes ablation studies on the two main novelties of constructing distant supervision signals. Without speaker masking in \S\ref{sec:masking} that encourages contextual reasoning, \ours{}$_{\text{en}}$ (-masking) drops $3.6\%$ on Emma. The use of pronoun-based coreference resolution described in \S\ref{sec:main_heuristics} also contributes to the final performance, as we see \ours{}$_{\text{en}}$ (-coref) drops accuracy on both English benchmarks. The other rules are from previous work so we do not repeat ablations for them in this paper.

We also manually analyze the distant supervision based on $100$ random extractions from \S\ref{sec:distant-supervision}. The accuracy of \ruleie~on these samples is $68\%$. We also find that $29\%$ require contextual reasoning as no direct evidence exists.  This, to some extent, explains the large gain achieved by our method on the implicit instances (Table~\ref{tab:chinese-category}). See examples of \ruleie{}'s output and analysis in Appendix~\ref{sec:appendix:anal}.




\section{Conclusion}
We propose a multi-step SI framework that includes \textbf{i)} \ruleie{}, a ruled-based system that is used to extract \textbf{ii)} 55K instances on English as distant supervision. We use them to train \textbf{iii)} a cross-lingual model \ours{} that outperforms previous systems on both English and Chinese benchmarks, by as much as $6.2\%$, and over $40\%$ in few-shot settings. Besides, our model works on cross-language SI even without in-domain or in-language supervision, achieving $89.5\%$ on the Chinese JY dataset with only English-based expert knowledge.



\section{Limitations}

Following some previous studies, we rely on off-the-shelf tools such as NER, coreference resolution, and SRL for generating or inferring candidate speakers, which may lead to error propagation or miss non-entity speakers who are not named by the authors. Relying on NLP tools may also create distribution biases, so that our model may favor certain types of utterances over the others. Method-wise, our model takes only certain length of the context into considerations, therefore global information such as character relationships might be missing. This might be one of the main source of errors. To this end, we do not include a thorough error analysis due to space limitations of a short paper, but we will release all resources and encourage future work to make progress on this front.

\bibliography{anthology,custom}
\bibliographystyle{acl_natbib}

\clearpage
\newpage
\appendix

\section{Appendix}
\label{sec:appendix}

\subsection{Data Statistics}
\label{sec:appendix:data}

\begin{table}[h!]
\centering
\footnotesize
\begin{tabular}{lllll}
\toprule
     & \textit{P\&P}    & \textit{Emma} & \textit{JY} & \textit{WP} \\
\midrule
language     & en & en  & zh & zh   \\
\# of training data    & 905 & --     & 17,159   & 2,000  \\
\# of dev data    & -- & --     & 5,719   & 298  \\
\# of test data    & 128 &  362    & 5,719   & 298 \\
\# of unique characters    & 52 &  49    & 94   & 125 \\
\bottomrule
\end{tabular}
\caption{Data statistics of SI datasets (en: English, zh: Chinese).}
\label{tab:data}
\end{table}

\subsection{Result Analysis}
\label{sec:appendix:anal}

We notice that \ours{}$_{\text{x}}$ obtain better performance on Chinese SI datasets versus the English datasets. One possible reason may be the changing writing styles over the years: the Chinese source novels such as WP were published more recently than the English novels P\&P and Emma published hundreds of years ago. We find that authors of these Chinese novels tend to write shorter utterances (e.g., 31.24 in the WP vs. 50.24 in P\&P), thus it may be easier for a span selection model to identify the direct speakers in nearby contexts considering the input length constraint of the backbone models.

In the following example in Table~\ref{table:extraction_example}, the speaker of the utterance \emph{``I wasn't far...been there.''} is correctly identified (Person X).


\begin{table}[h!]
\centering
\scriptsize
\begin{tabular}{p{1cm}p{6cm}}
\toprule
document & \emph{People: Person Q; Person D; Person G; Person T; Person N; Person Y; Person X; Person M; Person J; Person Z; Person K; Person C; Person F; Person L; Person B}\\\\
   & \emph{But to this Person L would not assent, alleging that he should feel himself to be a burden both to Lord and Person B.} \\\\
   & \emph{On the Person D did not go out, saying that he would avoid the expense, and on that day there was a good run.}\\\\
   & \emph{``It is always the way,'' said \textbf{Person X}. ``If you miss a day, it is sure to be the best thing of the season. An hour and a quarter with hardly anything you could call a check! It is the only very good thing I have seen since I have been here. Person T was with them all through.''}\\\\
   & \emph{``And I suppose you were with Person T. ''} \\\\
      & \emph{\underline{``I wasn't far off. I wish you had been there. ''} On the next day the meet was at the kennels, close to Person C, and someone drove his friend over in a gig....} \\\\
     
\midrule
question & who said \emph{``I wasn't far off. I wish you had been there. ''?} \\
speaker & Person X \\
\bottomrule
\end{tabular}
 \caption{A weakly-labeled instance example to train \ours{}.}
 \label{table:extraction_example}
\end{table}

\subsection{Implementation Details II}
Due to the input length constraints of the backbone models, we shorten the lengthy inputs in all SI datasets by removing the last sentences from \keywordCode{context} until the sequence fits the model.

\end{document}